\documentclass[twoside]{article}

\usepackage{graphicx}
\usepackage{svg}

\usepackage{amssymb, amsthm, bm, amsmath}

\newcommand\independent{\protect\mathpalette{\protect\independenT}{\perp}}
\def\independenT#1#2{\mathrel{\rlap{$#1#2$}\mkern2mu{#1#2}}}

\makeatletter
\def\thm@space@setup{%
  \thm@preskip=10pt 
  \thm@postskip=10pt 
}
\makeatother

\newtheorem{theorem}{Theorem}[section]
\newtheorem{proposition}{Proposition}[section]

\newtheorem{lemma}{Lemma}[section]

\theoremstyle{definition}
\newtheorem{definition}{Definition}[section]

\theoremstyle{definition}

\theoremstyle{definition}
\newtheorem{example}{Example}[section]

\theoremstyle{remark}

\usepackage{hyperref}
\usepackage[capitalise]{cleveref}
\usepackage[accepted]{aistats2025}
\usepackage[round]{natbib}

\usepackage[normalem]{ulem} 
\usepackage{xcolor}         

\begin{document}

%

%

\twocolumn[

\aistatstitle{On the Identifiability of Causal Abstractions}

\aistatsauthor{ Xiusi Li \And  Sékou-Oumar Kaba \And Siamak Ravanbakhsh }

\aistatsaddress{ Mila, McGill University \And  Mila, McGill University \And Mila, McGill University } ]

\begin{abstract}
   Causal representation learning (CRL) enhances machine learning models' robustness and generalizability by learning structural causal models associated with data-generating processes. We focus on a family of CRL methods that uses contrastive data pairs in the observable space, generated before and after a random, unknown intervention, to identify the latent causal model. \citep{brehmer2022weakly} showed that this is indeed possible, given that all latent variables can be intervened on \emph{individually}. However, this is a highly restrictive assumption in many systems. In this work, we instead assume interventions on \emph{arbitrary subsets} of latent variables, which is more realistic. 
   We introduce a theoretical framework that calculates the \emph{degree} to which we can identify a causal model, given a set of possible interventions, up to an \emph{abstraction} that describes the system at a higher level of granularity. 
\end{abstract}

\section{INTRODUCTION}

Causal representation learning (CRL) \citep{scholkopf2021toward} generalizes non-linear independent component analysis \citep{hyvarinen1999nonlinear, hyvarinen2019nonlinear} and causal discovery \citep{spirtes2001causation}, aiming to extract both latent variables and their causal graph in the form of structural causal models (SCMs). The question of identifiability naturally arises since we would like to guarantee that the set of models consistent with the given observable distribution is unique up to some equivalence class. In particular, we would like to know the level of granularity at which the true latent variables and the causal graph can be recovered.

We consider the problem of CRL assuming we have access to counterfactual data pairs \(\mathbf{x}\) and \(\Tilde{\mathbf{x}}\) from the same observable space, before and after a random, unknown intervention. This is necessitated by the infeasibility of learning disentangled representations from unsupervised observational data alone \citep{hyvarinen1999nonlinear, locatello2019challenging}, and is sometimes referred to as self-supervised \citep{von2021self}, contrastive \citep{zimmermann2021contrastive}, or weakly supervised learning \citep{shu2019weakly, locatello2020weakly, brehmer2022weakly, ahuja2022weakly}.

Previously, in this particular setting, it has been shown that the full causal graph can be recovered up to isomorphism \citep{brehmer2022weakly}, and all the latent variables can be recovered up to element-wise diffeomorphism. However, as was noted by the authors, this result relies on overly restrictive assumptions; for example, the number of nodes on the causal graph must be known in advance, and each node must be intervened upon individually with nonzero probability. 

Instead, we show that when we remove these assumptions, we can still identify ``coarser" versions of the causal model, known as its \emph{abstractions}, the related works of which we will expand on in the following paragraph. This relaxation is significant as it makes the setting much more realistic; in many systems, intervening on every variable individually is infeasible.

Causal abstraction \citep{rubenstein2017causal, beckers2019abstracting, rischel2020category, rischel2021compositional, otsuka2022equivalence, anand2023causal, massidda2023causal} is the study of how microscopic variables and causal mechanisms can be aggregated to macroscopic abstractions on a higher level while maintaining a notion of interventional consistency, which enables more efficient reasoning and interpretability \citep{geiger2021causal, geiger2024causal}. Since this is still an emerging avenue of research, a standard unified formalism has yet to be established. However, as was highlighted previously in \citep{zennaro2022abstraction}, all of the various definitions of causal abstractions, in one way or another, tend to have a \emph{structural} map dealing with the causal graph, and a \emph{distributional} map dealing with the variables and stochastic causal mechanisms associated with its nodes and edges respectively.

In classic causal inference, the idea of structural abstraction can in some sense already be found in the form of Markov equivalence classes, or interventional Markov equivalence classes \citep{Verma1990EquivalenceAS, hauser2012characterization, pmlr-v80-yang18a}, where in the simplest cases, directed edges on the causal graph are abstracted away to undirected ones. 

In CRL, the concept of distributional abstraction is more prevalent. For example, \citep{von2021self} introduces a definition for \emph{block-identifiability}, in which the grouping of latent variables into blocks essentially serves as an abstraction of individual latent variables. More recent works \citep{ahuja2022weakly, yao2023multi} also examine overlapping blocks of latent variables and the identifiability of their intersections, complements, and unions. 

In this paper, we investigate the identifiability of latent causal models up to their abstractions. Previous works in this setting have either focused on the problem of whether a pre-conceived latent causal model can be identified at all \citep{brehmer2022weakly}, or are only concerned with identifying abstractions of the latent variables without identifying abstractions of the latent causal graph \citep{von2021self, ahuja2022weakly, yao2023multi}.

To the best of our knowledge, we provide the first identifiability results that give a graphical criterion for the degree of abstraction which we can identify latent causal models up to, depending on the interventional data available within the context of the weakly-supervised CRL problem, which take into account both structural and distributional properties of SCMs.

We structure the remainder of this work as follows. In \cref{subsec: dgp} we introduce the weakly-supervised CRL problem setup in terms of the data generating process, which is essentially the same as the one outlined in \citep{brehmer2022weakly}. In \cref{subsec: id up to eq}, we proceed to give increasingly restrictive definitions of the identifiability of causal model parameters \emph{up to equivalence}, and in \cref{subsec: id up to abst} we move on to increasingly restrictive definitions of the identifiability of causal model parameters \emph{up to abstraction}. In \cref{sec: 2} we explain the assumptions behind our main results, before presenting the statements. We leave the detailed proofs of these results to the appendix but draw attention to some key properties of the data generating process in \cref{subsec: proof outlines}, and the proof techniques used at a high level.  In \cref{subsec: discussion} we provide some intuition for some of the unexpected conclusions that can be drawn from the results. Finally, in \cref{sec: applications}, we outline the various downstream applications of our results, as well their limitations.

\section{PROBLEM FORMULATION}
\label{sec: problem setup}

\subsection{DATA GENERATING PROCESS}
\label{subsec: dgp}

In this section, we describe the data-generating process in our problem setting, which is the functional relationship between the latent causal model parameters \(\theta\) and the resultant distribution \(p_\theta(\mathbf{x},  \Tilde{\mathbf{x}}) \) of counterfactual or contrastive pairs of observational data.

We first introduce the \emph{structural causal model} (SCM) describing the pre-intervention latent variables. Let \(\mathcal{G}\) be a directed acyclic graph, and associate each of the nodes \(i \in V(\mathcal{G})\) with a vector space \(\mathcal{Z}_i\), a random variable \(\mathbf{z}_i\) taking values on  \(\mathcal{Z}_i\), as well as a conditional probability distribution \(p \left( \mathbf{z}_i \mid \mathbf{z}_{Pa_{\mathcal{G}}(i)} \right)\). Furthermore, let each of the conditional distributions have a functional representation 
\footnote{The functional characterization of causal mechanisms, as was first introduced in \citep{Verma1990EquivalenceAS} allows us to define the effect of \emph{interventions} on the model, and is sometimes referred to as \emph{noise outsourcing} in contemporary literature (see e.g. \citet{bloem2020probabilistic}).}
\begin{equation}
\label{sem-z1}
    \mathbf{z}_i = f_i \left( \mathbf{z}_{Pa_{\mathcal{G}}(i)}, \bm \varepsilon_i \right), \quad \bm \varepsilon_i \sim p_{\bm \varepsilon_i} \quad \forall i \in V(\mathcal{G}),
\end{equation}
where the distributions of the \emph{exogenous variables} \(\bm \varepsilon_i \in \mathcal{E}_i\) are all mutually independent, and each \(f_i: \mathcal{Z}_{Pa_{\mathcal{G}}(i)} \times \mathcal{E}_i \to \mathcal{Z}_i\) is a deterministic function. Then we can denote \(\mathcal{Z}:= \bigoplus_{i \in V(\mathcal{G})} \mathcal{Z}_i\) and \(\mathcal{E}:= \bigoplus_{i \in V(\mathcal{G})} \mathcal{E}_i\), and define a deterministic function \(\mathbf{f}: \mathcal{E} \to \mathcal{Z}\) by successively applying the causal mechanisms \(f_i\). Therefore the distribution of the pre-intervention latents \(p(\mathbf{z}) \) can be parametrized by 
\begin{equation}
    \mathbf{\theta}_{\text{SCM}} := \left(\mathcal{G}, \mathbf{f}, p_{\bm \varepsilon} \right).
\end{equation}
We next describe the interventions on the latent variables. Let \(\bm\iota\) be a random variable taking values in the power set of vertices of \(\mathcal{G}\) (assumed to be finite), which tells us which latent variables are intervened upon at any one time.  We denote the distribution of this discrete random variable as \(P_{\bm \iota}\). Futhermore we write \(\mathcal{I} := \text{supp}(\bm \iota)\) and refer to its elements as \emph{intervention targets} (i.e. the subsets of nodes which get intervened upon with nonzero probability). In the event that \(\bm \iota = S\) for some \(S \subseteq V(\mathcal{G})\), we assume that for every node \(i\) in \(S\) the causal mechanism from \(Pa_{\mathcal{G}}(i)\) to \(i\) becomes completely severed, in what is known as a \emph{perfect} intervention. Therefore, the post-intervention latents \(\Tilde{\mathbf{z}} \) conditional on \(\bm \iota = S\) satisfy
\begin{equation}
\label{sem-z2i}
    \Tilde{\mathbf{z}}_i = \Tilde{f}^{(S)}_i(\bm \Tilde{\bm \varepsilon}_i) \quad \forall i \in S, \quad \Tilde{\bm \varepsilon}_S \sim p_{\Tilde{\bm \varepsilon}_S},
\end{equation}
\begin{equation}
\label{sem-z2j}
    \Tilde{\mathbf{z}}_j  = f_j \left( \Tilde{\mathbf{z}}_{Pa_{\mathcal{G}}(j)}, \bm \varepsilon_j \right) \quad \forall j \in V(\mathcal{G}) \setminus S,
\end{equation}
where the new exogenous variable \(\Tilde{\bm \varepsilon}_S\) corresponding to the target set post-intervention is independent from the pre-intervention exogenous variable \({\bm \varepsilon}\), and each \(\Tilde{f}^{(S)}_i\) is a new function. Thus we can describe the effect of a random, unknown intervention on the structural causal model. Therefore the joint distribution of pre-intervention and post-intervention latent pairs \(p(\mathbf{z}, \Tilde{\mathbf{z}})\) can be parametrized by \((\mathbf{\theta}_{\text{SCM}}, \mathbf{\theta}_{\text{intv}})\), where
\begin{equation}
    \mathbf{\theta}_{\text{intv}} := (\Tilde{\mathbf{f}}, p_{\Tilde{\bm \varepsilon}}, P_{\bm \iota} ).
\end{equation}
Finally, let \(\mathcal{X}\) be the vector space known as the observation space, and let \(g : \mathcal{Z} \to \mathcal{X}\) be an invertible \emph{mixing function} such that
\begin{equation}
\label{mixing}
    \mathbf{x} = g(\mathbf{z}), \quad \Tilde{\mathbf{x}} = g(\Tilde{\mathbf{z}}).
\end{equation}

Therefore the joint distribution of counterfactual pairs of observational data \(p(\mathbf{x},  \Tilde{\mathbf{x}})\) can be parametrized by 
\begin{equation}
    \theta := \left(\theta_{\text{SCM}}, \theta_{\text{intv}} , g \right).
\end{equation}

\begin{figure}[h]
    \centering
    \includegraphics[width=0.95\linewidth]{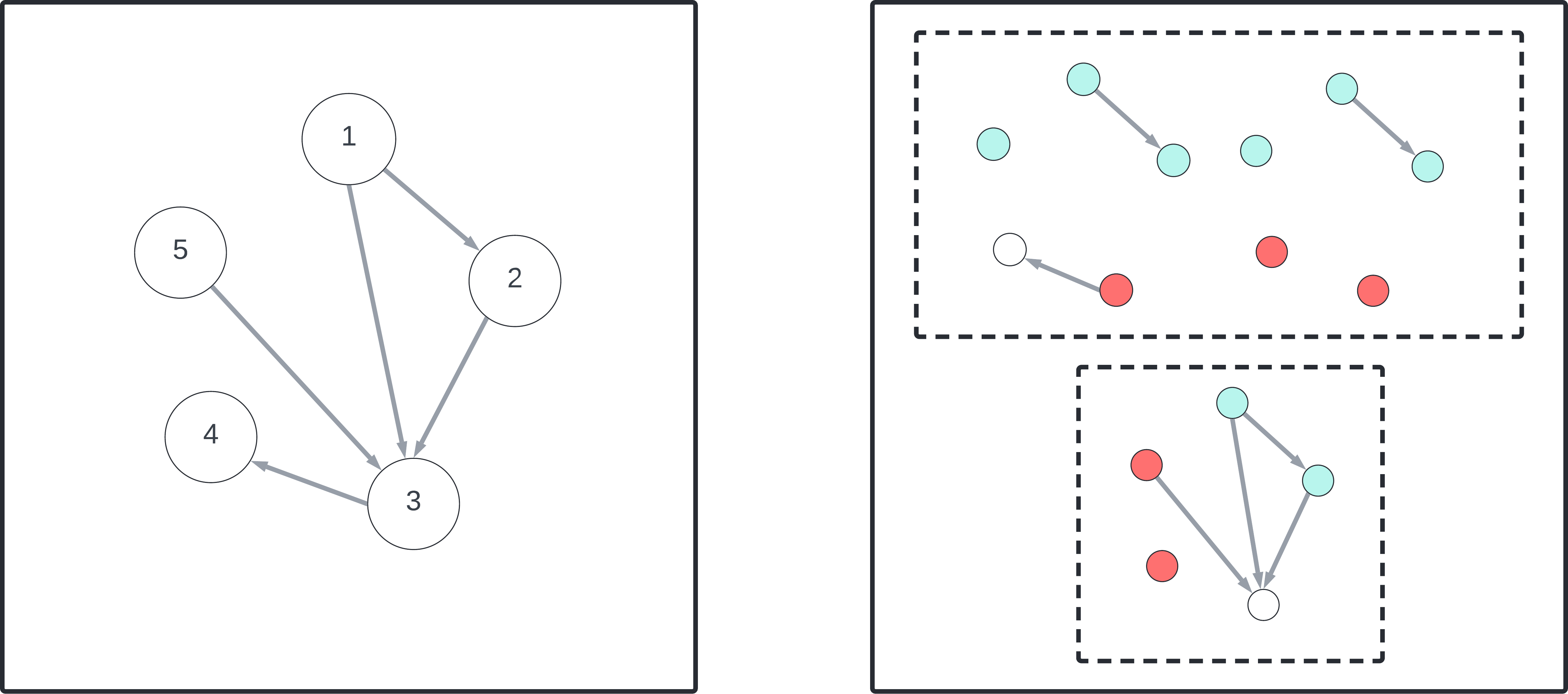}
    \caption{Data Generating Process for a Latent Causal Model. On the left we have the unintervened structural causal model generating samples of the pre-intervention latent \(\mathbf{z}\), while on the right we have a mixture of intervened structural causal models generating samples of the post-intervention latent \(\tilde{\mathbf{z}}\), for a collection of intervention targets \(\mathcal{I} = \{\{3\}, \{3, 4\}, \{4, 5\} \} \), as denoted by the subsets of red vertices; the corresponding non-descendant sets of the intervention targets are denoted by the subsets of blue vertices.}
    \label{fig:dgp}
\end{figure}

It is important to note that the counterfactual setting outlined here differs from methods using \emph{interventional data} \citep{brouillard2020differentiable, gresele2020incomplete, ahuja2023interventional, von2024nonparametric, zhang2024identifiability} in two respects.

\begin{enumerate}
    \item The counterfactual setting makes the more restrictive assumption of having access to the \emph{joint} distribution of \((\mathbf{x}, \Tilde{\mathbf{x}})\), whereas the interventional setting usually only assumes access to the marginal distributions of \(\mathbf{x}\) and \(\Tilde{\mathbf{x}}\) separately.
    \item The interventional setting makes the more restrictive assumption of being able to observe the \emph{type} of intervention that occurs, even though the exact intervention target corresponding to a given type may be unknown. Concretely, in the interventional setting, we assume access to multiple marginal distributions of \(\mathbf{x}^{(e)}\) indexed by an environment\footnote{This is also sometimes referred to as a ``view" \cite{yao2023multi}} variable \(e\). The crucial point here is that \(e\) is \emph{observable}, and that for any fixed \(e\), the latent variable \(\mathbf{z}^{(e)}\) is generated by a causal model with an \emph{invariant causal structure}. In comparison, in the counterfactual setting, the intervention variable \(\bm \iota\) is not observable, and samples of \(\Tilde{\mathbf{z}}\) are generated by a \emph{mixture} of causal models with distinct graphical structures.
\end{enumerate}

While CRL methods using interventional data are arguably more practical for applications such as biology \citep{belyaeva2021dci}, where we have access to data generated from known experimental settings; methods using counterfactuals are better suited to cases such as temporal data from dynamical systems or offline reinforcement learning \citep{lippe2022citris, ahuja2022weakly, brehmer2022weakly}, where at any given time step a \emph{random} intervention may take place. Further justification for the assumption of availability of counterfactual data pairs will be elaborated on in \cref{sec: applications}.

\subsection{IDENTIFIABILITY UP TO EQUIVALENCE}
\label{subsec: id up to eq}

\emph{Identifiability} in statistics refers to the ability to uniquely determine the true values of model parameters from observed data. A model is considered strongly identifiable if there is only a single set of parameter values that can generate the observed data. More generally, identifiability up to an equivalence class means that while there may be multiple sets of parameter values that can generate the same observed data, these sets are equivalent in some meaningful way.

For our purposes, we assume there exist some ground truth parameters \({\theta}^\star\), and that we may take unlimited samples from the latent causal model \(p_{\theta^\star}(\mathbf{x}, \Tilde{\mathbf{x}})\), in order to learn an estimator of \(\theta^\star\). We say that \(\theta^\star\) is identifiable up to some equivalence relation \(\sim\) with respect to some hypothesis class of model parameters \(\Theta\), if for any \(\theta \in \Theta\)
\begin{equation}
\label{eq: identifiabilty}
    p_{\theta}(\mathbf{x},  \Tilde{\mathbf{x}}) = p_{\theta^\star}(\mathbf{x}, \Tilde{\mathbf{x}}) \implies   \theta \sim \theta^\star.
\end{equation}

\subsubsection{LATENT DISENTANGLEMENT}

One instance of an equivalence relation between \(\theta^\star\) and \(\theta\) is defined in terms of an equivalence relation between the corresponding mixing function \(g^\star\) and \(g\). It is known as \emph{disentanglement} \citep{bengio2013representation,higgins2018towards}, and we will draw attention to two variants of its definition.

The first variant deals with a \emph{single latent subspace} that we want to isolate, which loosely speaking means that we want any two equivalent models to produce latent distributions that have the same marginals when restricted to this subspace.

\begin{definition}[Latent disentanglement]
\label{def: ld}
    Given latent causal models parametrized by  \(\theta^\star\) and \(\theta\) as defined in \cref{subsec: dgp}, and latent subspaces \(\mathcal{W}^\star \subset \mathcal{Z}^\star\) and \(\mathcal{W} \subset \mathcal{Z}\), we say that
    \[\theta \sim_{\text{L}} \theta^\star \]
     with respect to these subspaces, if there exists a measurable function
     \(
         h: \mathcal{W}^\star \to \mathcal{W}
     \)
     such that if we define the random variables \(\mathbf{w}^\star\) and \(\mathbf{w}\) to be the latent components of \(\mathbf{z}^\star\) and  \(\mathbf{z}\) 
    corresponding to the latent subspace \(\mathcal{W}^\star\) and \(\mathcal{W}\) respectively 
    , then the function \(h\) satisfies 
    \begin{equation}
        \mathbf{w} \overset{d}{=} h(\mathbf{w}^\star)
    \end{equation}
\end{definition}

Alternatively, we can say that the \emph{encoder} \(g^{-1}: \mathcal{X} \to \mathcal{Z}\) disentangles the ground truth latent variable \(\mathbf{w}^\star\) by identifying it with the variable \(\mathbf{w}\) in the resultant latent representation \(\mathbf{z}\).

We emphasize that the above definition is a \emph{distributional equivalence between a single pair of latent components} in two causal models, without placing constraints on any of the other components.

The second variant of the definition of disentanglement deals with a full decomposition of the latent space into a \emph{direct sum of latent subspaces}, such that we want any two equivalent models to produce latent distributions that have the same marginals when restricted to any of the subspaces in this decomposition.

\begin{definition}[Full latent disentanglement]
    Given latent causal models parametrized by  \(\theta^\star\) and \(\theta\) as defined in \cref{subsec: dgp}, and latent space decompositions \(\mathcal{Z}^\star = \bigoplus_{i = 1}^n \mathcal{Z}^\star_i\) and \(\mathcal{Z} = \bigoplus_{j = 1}^n \mathcal{Z}_j\), we say that 
    \[\theta \sim_{\text{FL}} \theta^\star\] 
    with respect to these decompositions, if there exist a \emph{bijective function} \(\phi :   \{1, \cdots, n\} \to \{1, \cdots, n\}\)
    and measurable functions 
    \(
        h_i : \mathcal{Z}^\star_{i} \to \mathcal{Z}_{\phi(i)} 
    \)
    for all \(i \in \{1, \cdots, n\}\) such that if we define the random variables \(\mathbf{z}^\star_i\) and \(\mathbf{z}_j\) to be the latent components of \(\mathbf{z}^\star\) and \(\mathbf{z}\)
    corresponding to the latent subspaces \(\mathcal{Z}^\star_i\) and \(\mathcal{Z}_j\) respectively, then for all \(i \in \{1, \cdots, n\}\) we have
    \begin{equation}
    \label{eq: latent disentanglement}
        \mathbf{z}_{\phi(i)} \overset{d}{=} h_{i}\left(\mathbf{z}^\star_{i}\right)
    \end{equation}
\end{definition}

Alternatively, we can say that the encoder \(g^{-1}\) produces a fully disentangled representation \( (\mathbf{z}_{1}, \cdots, \mathbf{z}_{n})\)  of the ground truth latent variables \( 
 (\mathbf{z}^\star_1, \cdots, \mathbf{z}^\star_n)\).

We emphasize that the above definition consists of \emph{distributional equivalences between the full sets of latent components} in two causal models, up to the permutation given by \(\phi\), therefore it is easy to check that the following statement holds.

\begin{lemma}
    Given latent causal models parametrized by  \(\theta^\star\) and \(\theta\) as defined in \cref{subsec: dgp}, suppose \(\theta \sim_{\text{FL}} \theta^\star\) with respect to the decompositions \(\bigoplus_{j = 1}^n \mathcal{Z}_j\)  and \(\bigoplus_{i = 1}^n \mathcal{Z}^\star_i \).  Then for all \(i \in \{1, \cdots, n\}\), we have \(\theta \sim_{\text{L}} \theta^\star\) with respect to \(\mathcal{Z}_{\phi(i)}\) and \(\mathcal{Z}^\star_i\).
\end{lemma}

\subsubsection{STRUCTURAL CAUSAL MODEL ISOMORPHISM}

Note that so far, we have only defined distributional equivalences between the variables of the latent causal model and their representations. An even stronger equivalence relation than full latent disentanglement between \(\theta^\star\) and \(\theta\) takes causal structure into account, and requires that in addition the causal graphs \(\mathcal{G}^\star\) and \(\mathcal{G}\) be \emph{isomorphic}. This is sometimes referred to as \emph{structural causal model isomorphism} \citep{fong2013causal, brehmer2022weakly} or the CRL identifiability class \cite{von2024nonparametric}.

\begin{definition}[SCM isomorphism]
        Given latent causal models parametrized by  \(\theta^\star\) and \(\theta\) as defined in \cref{subsec: dgp}, note that there exist \emph{canonical decompositions} of their latent subspaces \(\mathcal{Z}^\star\) and \(\mathcal{Z}\) into direct sums of subspaces \(\bigoplus_{i \in V(\mathcal{G}^\star)} \mathcal{Z}_i^\star \) and \(\bigoplus_{j \in V(\mathcal{G})} \mathcal{Z}_j\) respectively. We say that
    \[\theta \sim_{\text{SCM}} \theta^\star\]
    if there exists a \emph{graph isomorphism} \(\phi: \mathcal{G}^\star \to \mathcal{G}\), and measurable functions \( h_i : \mathcal{Z}^\star_{i} \to \mathcal{Z}_{\phi(i)}\)
   such that Eq\eqref{eq: latent disentanglement} holds for all \(i \in V(\mathcal{G}^\star)\).
    
    Alternatively, we say that the structural causal models with parameters \(\theta_{\text{SCM}}\) and \(\theta_{\text{SCM}}^\star\) are \emph{isomorphic}.

Note that the above definition consists of a \emph{structural equivalence} in the form of the graph isomorphism, as well as distributional equivalences between latent components in two causal models that are compatible with the graph isomorphism, therefore it is easy to check that the following statement holds. 

\begin{lemma}
    Given latent causal models parametrized by  \(\theta^\star\) and \(\theta\) as defined in \cref{subsec: dgp}, suppose \(\theta \sim_{\text{SCM}} \theta^\star\). Then \(\theta \sim_{\text{FL}} \theta^\star\) with respect to \(\bigoplus_{j \in V(\mathcal{G})} \mathcal{Z}_j \) and \(\bigoplus_{i \in V(\mathcal{G}^\star)} \mathcal{Z}_i^\star \).
\end{lemma}

\end{definition}

\subsection{IDENTIFIABILITY UP TO ABSTRACTION}
\label{subsec: id up to abst}

In this section, we introduce the concept of identifiability up to model abstraction, which can be thought of as the ``common factor" between all models which are consistent with the observable distribution. 

To do this, we need some notion of a \emph{partial order} \(\preceq\) on \(\Theta\), which compares causal models by their level of granularity or, in some sense, complexity. Broadly speaking, given any causal model \(\theta^\star\), there is always another, more complex model \(\theta' \succeq \theta^\star\) that produces the same observable distribution
\footnote{For example, we can always add more ``dummy variables" that increase the number of nodes on the causal graph, which, when marginalized upon do not produce any effect on the final observable distributions under intervention}. 
What we want to know is what all the models which are able to produce the same observable distribution as \(\theta^\star\) have in common, meaning that we want to find an abstraction \({\theta}\) of \(\theta^\star\) that is the \emph{infimum} of all these models. Note that this principle of finding minimal causal structures consistent with the observed data, which can be thought of as a reformulation of Occam's razor, is discussed at length in Chapter 2 of \cite{pearl2009causality}.

Concretely, in our problem setting, \(\theta^\star\) is said to be identifiable up to the abstraction \({\theta}\) with respect to \(\preceq\) and some hypothesis class of parameters \(\Theta\) if for all \(\theta' \in \Theta\)
 \begin{equation}
    p_{\theta'}(\mathbf{x},  \Tilde{\mathbf{x}}) = p_{\theta^\star}(\mathbf{x}, \Tilde{\mathbf{x}}) \implies   {\theta} \preceq {\theta'}
\end{equation}

We will proceed to extend the definitions of equivalence relations \(\sim\) on the parameter space \(\Theta\) from the previous subsection to (weak) partial orders \(\preceq\) on \(\Theta\). 

\subsubsection{LATENT ABSTRACTION}

\begin{definition}[Latent abstraction]
    Given latent causal models parametrized by  \(\theta\) and \(\theta'\) as defined in \cref{subsec: dgp}, and a latent subspace \(\mathcal{W} \subset \mathcal{Z}\), together with a set of complementary latent subspaces \(\mathcal{Z}'_1, \cdots, \mathcal{Z}'_k \subseteq \mathcal{Z}'\), we say that 
    \[\theta \preceq_{\text{L}} \theta'\]
     with respect to this latent subspace in \(\mathcal{Z}\) and set of latent subspaces in \(\mathcal{Z}'\) if there exists measurable functions \(h_i: \mathcal{Z}_i' \to \mathcal{W}\) for all \(i \in \{1, \cdots, k\}\) such that if we define the random variables \(\mathbf{z}'_i\) to be the latent components of \(\mathbf{z}'\)
    corresponding to the latent subspaces \(\mathcal{Z}'_i\), then 
     \begin{equation}
         \mathbf{w} \overset{d}{=} \sum_{i = 1}^k  h_i(\mathbf{z}'_i).
     \end{equation}
\end{definition}

   Alternatively, we can say that the encoder \(g^{-1}\) produces an abstraction \(\mathbf{w}\) of the latent variables \((\mathbf{z}_1', \cdots, \mathbf{z}_k')\).

    Note that here we have defined a \emph{single distributional equivalence} between a subset of latent components in one model, and the distribution of a single latent component in another.

\begin{definition}[Full latent abstraction]
    Given latent causal models parametrized by  \(\theta\) and \(\theta'\) as defined in \cref{subsec: dgp}, and decompositions of the latent spaces \(\mathcal{Z}'\) and \(\mathcal{Z}\) into direct sums of subspaces \(\mathcal{Z}'_1 \oplus \cdots \oplus \mathcal{Z}'_n\) and \(\mathcal{Z}_1 \oplus \cdots \oplus \mathcal{Z}_m\) respectively, we say that 
    \[\theta \preceq_{\text{FL}} \theta'\] 
    with respect to these decompositions if there exist a \emph{surjective function} \(\phi :   \{1, \cdots, n\} \to \{1, \cdots, m\}\)
    and measurable functions \(h_i : \mathcal{Z}'_{i} \to \mathcal{Z}_{\phi(i)}\) such that if we define the random variables \(\mathbf{z}'_i\) and \(\mathbf{z}_j\) to be the latent components of \(\mathbf{z}'\) and \(\mathbf{z}\) 
    corresponding to the latent subspaces \(\mathcal{Z}'_i\) and \(\mathcal{Z}_j\) respectively, then for all \(j \in \{1, \cdots, m\}\), we have
    \begin{equation}
    \label{eq: latent abstraction}
        \mathbf{z}_{j} \overset{d}{=} \sum_{i \in \phi^{-1}(j)} h_i(\mathbf{z}^\star_i).
    \end{equation}
\end{definition}

Alternatively, we can say that the encoder \(g^{-1}\) produces an abstraction \((\mathbf{z}_1, \cdots, \mathbf{z}_m)\) of the latent variables \((\mathbf{z}'_1, \cdots, \mathbf{z}'_n)\)

    Note that here we have defined a \emph{full set of distributional equivalences} based on a surjection between all the latent components of two causal models, therefore it is easy to check that the following statement holds.

\begin{lemma}
    Given latent causal models parametrized by  \(\theta\) and \(\theta'\) as defined in \cref{subsec: dgp}, suppose \(\theta \preceq_{\text{FL}} \theta'\) with respect to the decompositions \(\mathcal{Z}'_1 \oplus \cdots \oplus \mathcal{Z}'_n\) and \(\mathcal{Z}_1 \oplus \cdots \oplus \mathcal{Z}_m\). Then for all \(j \in \{1, \cdots, m\}\), we have \(\theta \preceq_{\text{L}} \theta'\) with respect to the latent subspace \(\mathcal{Z}_j \subseteq \mathcal{Z}\) and the set of latent subspaces \(\{\mathcal{Z}'_i\}_{i \in \phi^{-1}(j)}\).
\end{lemma}

\subsubsection{STRUCTURAL CAUSAL MODEL HOMOMORPHISM}

To extend the definition of a structural causal model isomorphism, we make use of a more general definition of structure preserving maps between SCMs that is explored in \citep{otsuka2022equivalence}.

\begin{figure}[h]
    \centering
    \includegraphics[width=0.75\linewidth]{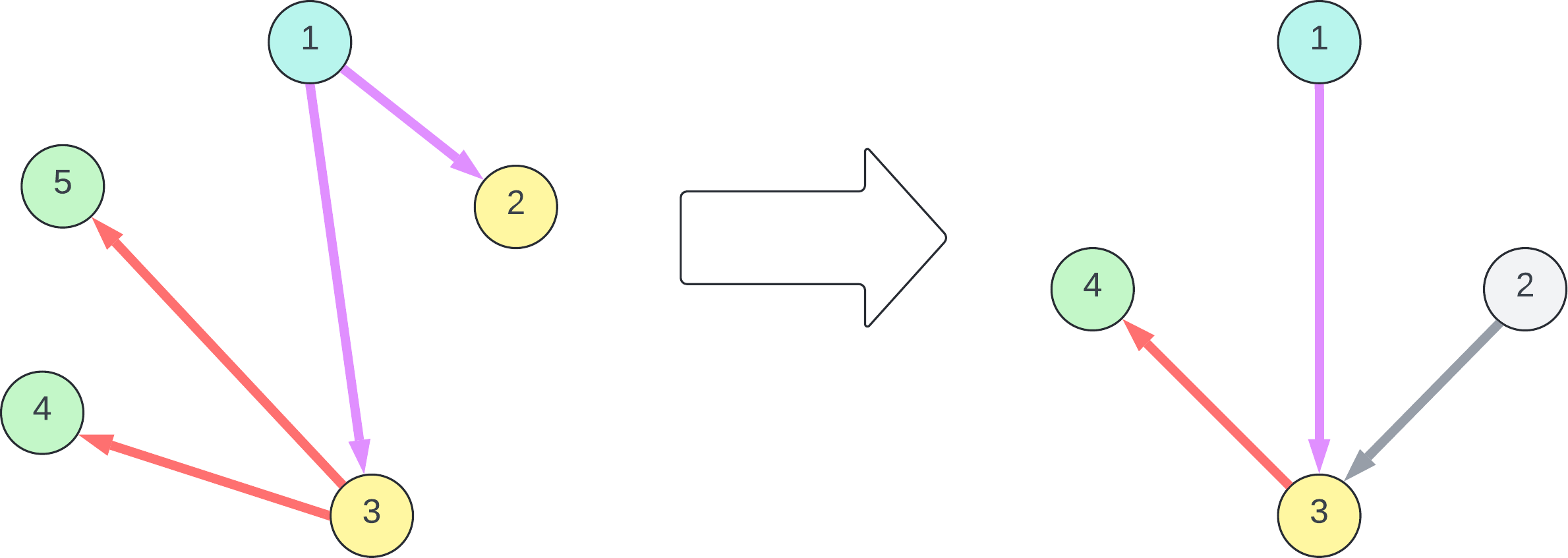}
    \caption{SCM homomorphism, as defined by a graph homomorphism $\phi$ that maps nodes in $\mathcal{G}'$ to nodes in $\mathcal{G}$ of the same colour (e.g. $\phi(2) = \phi(3) = 3$), as well as a set of invertible measurable functions which ensure that the latent variables represented by nodes in the image of $\phi$, which in this case are $\mathbf{z}'_1$, $\mathbf{z}'_3$ and $\mathbf{z}'_4$, have equivalent distributions to their counterparts after marginalization on $\mathbf{z}'_2$.} 
    \label{fig:causal-model-homomorphism}
\end{figure}

\begin{definition}[SCM homomorphism]
     Given latent causal models parametrized by  \(\theta\) and \(\theta'\) as defined in \cref{subsec: dgp}, we say that there exists a \emph{structural causal model homomorphism} between SCMs with parameters \(\theta'_{\text{SCM}}\) and \(\theta_{\text{SCM}}\)
    if there exists a \emph{graph homomorphism} \(\phi: \mathcal{G}' \to \mathcal{G}\), and measurable functions \(h_i : \mathcal{Z}'_{i} \to \mathcal{Z}_{\phi(i)}\) such that  Eq\eqref{eq: latent abstraction} holds for all \(j \in V(\mathcal{G})\). See \cref{fig:causal-model-homomorphism} for example.
\end{definition}

\begin{figure}[h]
    \centering
    \includegraphics[width=0.95\linewidth]{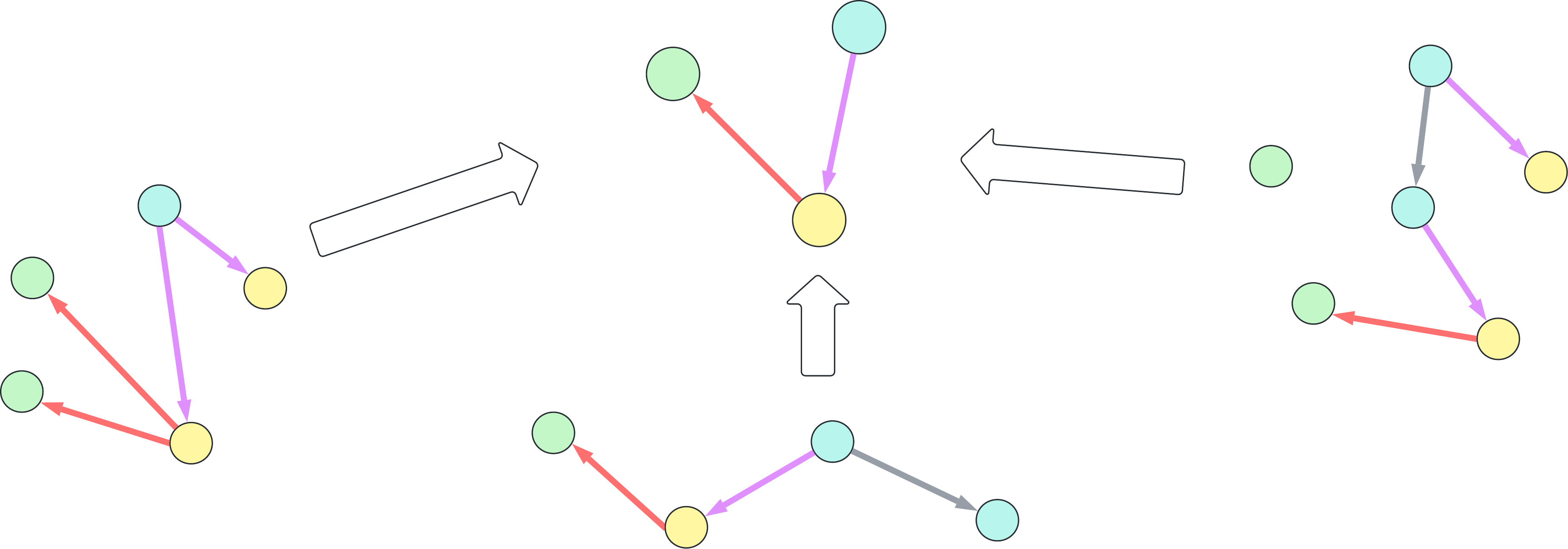}
    \caption{SCM Abstractions. The white arrows represent surjective SCM homomorphisms, which map causal models operating at higher granularity levels to lower granularity levels, such that the model at the centre is an abstraction of all the other models.}
    \label{fig:causal-model-abstraction}
\end{figure}

\begin{definition}[SCM abstraction]
\label{def: causal model abstraction}
   Given latent causal models parametrized by  \(\theta\) and \(\theta'\) as defined in \cref{subsec: dgp}, we say that \[\theta \preceq_{\text{SCM}} \theta'\] if there exists a SCM homomorphism between SCMs with parameters \(\theta'_{\text{SCM}}\) and \(\theta_{\text{SCM}}\), and the associated graph homomorphism \(\phi: \mathcal{G}' \to \mathcal{G}\) is surjective (i.e. an epimorphism). See \cref{fig:causal-model-abstraction} for example.
\end{definition}

Alternatively, we can say that \(\theta_{\text{SCM}}\) is an \emph{abstraction} of the SCM parameters \(\theta'_{\text{SCM}}\). 

Note that the above definition consists of a \emph{structural map} in the form of a graph epimorphism, as well as a \emph{full set of distributional equivalences} between all the latent components of two causal models that is compatible with the structural map, therefore it is easy to check that the following statement holds.

\begin{lemma}
    Given latent causal models parametrized by \(\theta\) and \(\theta'\) as defined in \cref{subsec: dgp}, suppose \(\theta \sim_{\text{SCM}} \theta'\).
    Then \(\theta \preceq_{\text{FL}} \theta'\) with respect to the canonical decompositions \(\bigoplus_{j \in V(\mathcal{G})} \mathcal{Z}_j \) and \(\bigoplus_{i \in V(\mathcal{G}')} \mathcal{Z}_i' \).
\end{lemma}

Finally, we can check that the equivalence relations from \cref{subsec: id up to eq} are consistent with the definitions of partial orders that give rise to the corresponding abstractions defined in this section.

\begin{lemma}
    Given latent causal models parametrized by \(\theta\) and \(\theta'\) as defined in \cref{subsec: dgp}, and any \((\sim, \preceq) \in \{(\sim_L, \preceq_L), (\sim_FL, \preceq_FL), (\sim_\text{SCM}, \preceq_\text{SCM})\}\)
    \[\theta \sim \theta' \iff  \theta \preceq \theta' \text{ and }  \theta \succeq \theta' \]
\end{lemma}

\section{IDENTIFIABILITY RESULTS}
\label{sec: 2}

\subsection{PRELIMINARIES}
Before presenting our main result, we will introduce several key concepts on which it is based, as well as some notation. We will assume that the reader is familiar with the definitions of $\sigma$-algebras and partitions, which can otherwise be found in \cref{app: defns}.

\paragraph{Family of non-descendants}
Given a directed graph \(\mathcal{G}\), the \emph{non-descendants} of a subset \(S\) of its vertices shall be denoted as \(\text{nd}(S)\), that is the intersection of the non-descendants of all the nodes in \(S\). Furthermore, for a family \(\mathcal{I}\) of intervention targets, we will denote the corresponding family of non-descendants as \(\mathbf{nd}(\mathcal{I}):= \{\text{nd}(S) : S \in \mathcal{I} \}\)

\paragraph{$\sigma$-algebra generated by family of sets}
Given a collection of subsets $\mathcal{A}$ of a set $X$, the \emph{$\sigma$-algebra generated by $\mathcal{A}$}, denoted $\sigma(\mathcal{A})$, is the smallest $\sigma$-algebra on $X$ that contains all the sets in $\mathcal{A}$. More formally, $\sigma(\mathcal{A})$ is the intersection of all $\sigma$-algebras on $X$ that contain $\mathcal{A}$, ensuring that $\sigma(\mathcal{A})$ satisfies the properties of a $\sigma$-algebra (containing the empty set, closed under complements and countable union).

\paragraph{Partition generated by $\sigma$-algebra}
The \emph{partition generated by a $\sigma$-algebra} $\mathcal{F}$ on a set $X$, denoted $\mathcal{P}({\mathcal{F}})$, is the collection of disjoint measurable sets in $\mathcal{F}$ that together cover $X$. More formally, it consists of the equivalence classes of the relation that considers two elements $x, y \in X$ to be equivalent if every set in $\mathcal{F}$ either contains both or contains neither. These equivalence classes form a partition, and each class is an element of the $\sigma$-algebra. This partition is maximal in the sense that the elements of the partition cannot be further subdivided using sets from $\mathcal{F}$.

\paragraph{Quotient graph generated by partition}
The \emph{quotient graph} $\mathcal{G} / \mathcal{P}$ of a directed graph $\mathcal{G}$ with respect to a partition $\mathcal{P} = \{A_1, A_2, \dots, A_k\}$ of the vertex set $V(\mathcal{G})$ is defined such that the vertex set of $\mathcal{G} / \mathcal{P}$ is exactly \(\mathcal{P}\), and for distinct blocks $A_i$ and $A_j$ in $\mathcal{P}$, there is a directed edge from $A_i$ to $A_j$ in $\mathcal{G} / \mathcal{P}$ if and only if there exists at least one directed edge from a vertex in $A_i$ to a vertex in $A_j$ in the original directed graph $\mathcal{G}$.

\paragraph{Graph condensations} Note that the definition above implies that given a directed acyclic graph \(\mathcal{G}\) and an arbitrary partition \(\mathcal{P}\) of its vertices, the resultant quotient graph \(\mathcal{G}' := \mathcal{G} / \mathcal{P}\) can contain cycles. However, it is always true that there exists another quotient graph \(\mathcal{G}'_C\), known as the \emph{condensation} of \(\mathcal{G}'\), that is acyclic. This is important because it ensures that we can always obtain an abstraction of some causal model for any partition of the vertices by taking the condensation.

We construct the condensation \(\mathcal{G}'_C\) by taking the quotient of \(\mathcal{G}'\) with respect to the partition defined by all the \emph{strongly
connected components} of \(\mathcal{G}'\), which are maximal subgraphs \( \mathcal{H} \subseteq \mathcal{G}' \) such that for all vertices \( u, v \in V(\mathcal{H}) \), there exists a directed path from \( u \) to \( v \) and from \( v \) to \( u \).

\subsection{MAIN RESULTS}

For both of the results stated in this section, we make the following assumptions on the hypothesis class of parameters \(\Theta\)

\begin{enumerate}
    \item \emph{Faithfulness of the causal graph}: Let \(\mathcal{G}\) be a perfect map \citep{pearl2009causality} for the distribution of \(p(\mathbf{z})\), meaning that it encapsulates all the conditional independences.
    \item \emph{Absolute continuity of latent distributions}: Let \(\mathcal{E}_i \cong \tilde{\mathcal{E}}_i \cong \mathcal{Z}_i\)\footnote{Here we use \(\cong\) to denote isomorphic vector spaces}, let \(f_i\), \(f_i^{(S)}\) be continuously differentiable for all \(i\) and \(S\), and let \(p_{\bm{\varepsilon}}\) and \(p_{\tilde{\bm{\varepsilon}}}\) be absolutely continuous.
    \item \emph{Smoothness of mixing function}:  Let \(g\) be a diffeomorphism.
\end{enumerate}

\subsubsection{IDENTIFIABLE MODEL ABSTRACTION}

Our first result shows that the parameters of a latent causal model as defined in \cref{subsec: dgp} can be identified up to  a SCM abstraction, depending on the \emph{non-descendant sets} of the intervention targets.

\begin{theorem}
\label{thm: 1}
Any latent causal model with parameters \(\theta^\star \in \Theta\) is identifiable up to a SCM abstraction \({\theta}\) with causal graph 
\begin{equation}
\label{eq: id quotient graph}
    {\mathcal{G}} = \mathcal{G}^\star / \mathcal{P}({\sigma( \mathbf{nd}(\mathcal{I^\star}))}).
\end{equation}
meaning that for all \(\theta' \in \Theta\) 
 \begin{equation}
    p_{\theta'}(\mathbf{x},  \Tilde{\mathbf{x}}) = p_{\theta^\star}(\mathbf{x}, \Tilde{\mathbf{x}}) \implies   {\theta} \preceq_{\text{SCM}} {\theta'}.
\end{equation}
Furthermore, we can show that the quotient graph \(\mathcal{G}\) is acyclic, so we do not have to resort to taking its condensation.
\end{theorem}

\begin{figure}[h]
    \centering
    \includegraphics[width=0.75\linewidth]{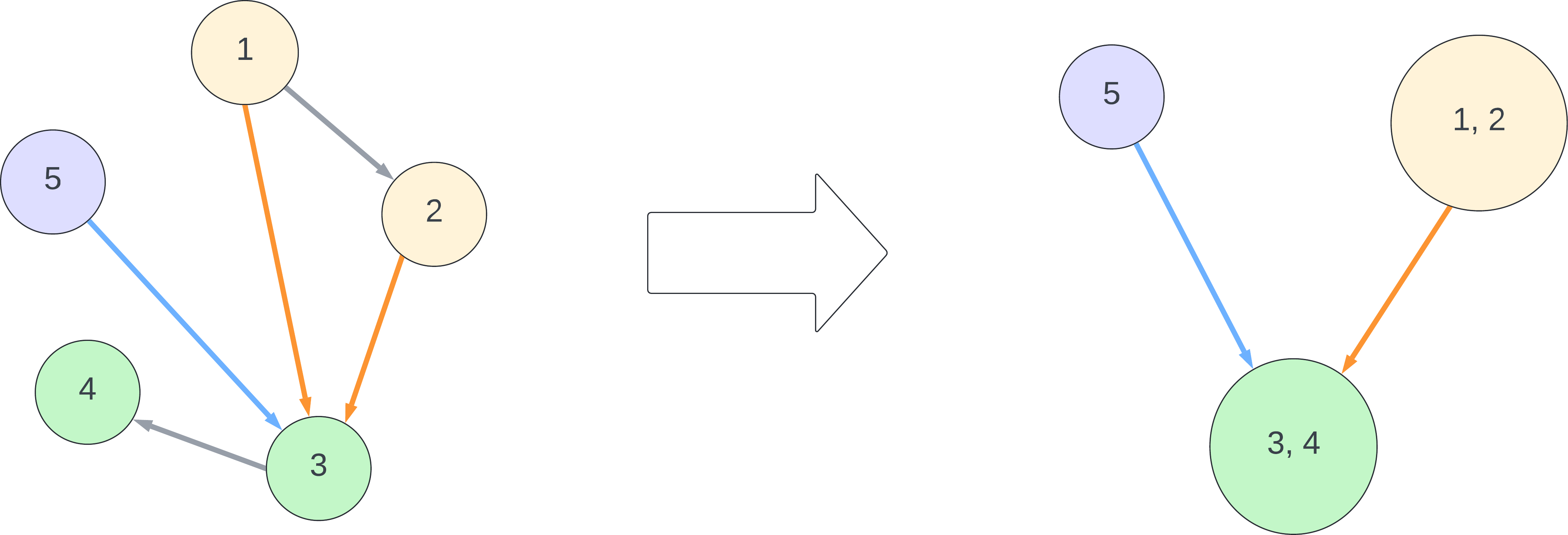}
    \caption{Identifiable SCM abstraction with graphical structure as shown on the left and $\mathcal{I}^\star = \{\{3\}, \{3, 4\}, \{4, 5\} \}$ as its family of  intervention targets.}
    \label{fig:nd thm}
\end{figure}

\begin{example}
    Let \(\theta^\star\) be the latent causal model depicted in \cref{fig:dgp}, so that \(\mathcal{I}^\star = \{\{3\}, \{3, 4\}, \{4, 5\} \}\). Then the corresponding family of non-descendants is \(\mathbf{nd}(\mathcal{I}^\star) = \{\{1, 2, 5\}, \{1, 2\} \}\), as shown by the subsets of blue vertices on the panel on the right of the figure. Hence we can compute the partition \(\mathcal{P}({\sigma( \mathbf{nd}(\mathcal{I}^\star))}) = \{\{1, 2\}, \{3, 4\}, \{5\}\}\), and know that we can identify the latent causal model up to the SCM abstraction shown in \cref{fig:nd thm}.
\end{example}

\subsubsection{ADDITIONAL IDENTIFIABLE LATENTS}

Our second result follows from the first, and uses our definition of latent disentanglement \(\sim_{\text{L}}\) to identify additional latents. Here we have a notion of equivalence between latent variables that correspond to the \emph{intersections} of intervention targets with the same non-descendant sets, but no constraints on the causal mechanisms between them.

\begin{theorem}
\label{thm: 2}
    For any \(N \in \mathbf{nd}(\mathcal{I}^\star)\) denote the intersection of all intervention targets with \(N\) as their non-descendant set as \[\pi(N) := \bigcap \{S \in \mathcal{I}: \text{nd}(S) = N\}.\] 
    Now provided that \(\pi(N)\) is a singleton set \(\{i\}\) and \(\mathcal{Z}^\star_{i} \cong \mathbb R\), then we can identify the latent \(\mathbf{z}^\star_{\pi(N)}\) up to disentanglement, meaning that for all \(\theta \in \Theta\)
    \begin{equation}
         p_{\theta}(\mathbf{x},  \Tilde{\mathbf{x}}) = p_{\theta^\star}(\mathbf{x}, \Tilde{\mathbf{x}}) \implies   {\theta} \sim_{L} {\theta^\star}
    \end{equation} 
    with respect to \(\mathcal{Z}^\star_{i}\) and some latent subspace in \(\mathcal{Z}\).
\end{theorem}

\begin{example}
        Let \(\theta^\star\) be the latent causal model depicted in \cref{fig:dgp}, so that \(\mathcal{I}^\star = \{\{3\}, \{3, 4\}, \{4, 5\} \}\) and \(\mathbf{nd}(\mathcal{I}^\star) = \{\{1, 2, 5\}, \{1, 2\} \}\) as before. Then by looking at the intersections of subsets of blue vertices in each of the boxes with dotted lines, we can see that \(\pi(\{1, 2\}) = \{4, 5\}\) and \(\pi(\{1, 2, 5\}) = \{3\}\). Thus we can disentangle the latent variable \(\tilde{\mathbf{z}}^\star_3\), provided that it takes value on \(\mathbb R\).
\end{example}

\subsection{PROOF OUTLINES}
\label{subsec: proof outlines}

We leave the precise details of the proofs of \cref{thm: 1} and \cref{thm: 2} to \cref{app: proofs}, and instead highlight some key properties of the data generating process that the proofs depend on. All together, these should serve as an outline of the main techniques used.

\paragraph{Finite mixtures} Since the intervention target \(\bm \iota\) is a discrete random variable and takes only a finite number of values, the latent distribution becomes a finite mixture of distributions with one mixture component for each value of $S \in \mathcal{I}$
\begin{equation}
\label{eq: finite-mixtures}
    p(\mathbf{z}, \Tilde{\mathbf{z}}) = \sum_{S \in \mathcal{I}} \mathbb{P}({\bm \iota} =S) p(\mathbf{z}, \Tilde{\mathbf{z}} \mid \bm \iota = S).
\end{equation}
We will show that these components can be separated up to equivalence classes of \(\mathcal{I}\) where each component corresponds to an element \( N \in \mathbf{nd}(\mathcal{I})\).
\begin{equation}
    p(\mathbf{z}, \Tilde{\mathbf{z}}) = \sum_{ N \in \mathbf{nd}(\mathcal{I})} \mathbb{P}(\text{nd}({\bm \iota}) = N) p(\mathbf{z}, \Tilde{\mathbf{z}} \mid \text{nd}({\bm \iota}) = N).
\end{equation}

\paragraph{Invariance of non-descendant variables}
From \cref{sem-z1} and \cref{sem-z2j}, we can see that given any intervention target \(S\), the block of latents which correspond to the non-descendant set of \(S\) (i.e. nodes in \(\mathcal{G}\) which are not descendants of any member of \(S\)), is invariant across the counterfactual pair, and crucially is the ``maximally" invariant block. Formally, if we let \(N = \text{nd}(S)\) denote the non-descendant set of \(S\) then
\begin{align}
    & \mathbb{P} (\mathbf{z}_{N} \neq \Tilde{\mathbf{z}}_{N} \mid \bm \iota = S) = 0,  \label{eq: invariant block}\\
    & \mathbb{P} (\mathbf{z}_{T} \neq \Tilde{\mathbf{z}}_{T} \mid \bm \iota = S) > 0 \quad \forall T \not \subseteq N. \label{eq: max invariant block}
\end{align}
\cite{von2021self} made use of this property to isolate the \emph{combined} ``content" block, \(\bigcap  \mathbf{nd}(\mathcal{I}^\star)\), from the ``style" block, which consists of the remainder of the latents, identifying the true causal model up to the abstraction \(\mathbf{z}_{\text{content}} \to \mathbf{z}_{\text{style}}\). However, through a slightly more careful examination, we show that every block in \( \mathcal{P}({\sigma( \mathbf{nd}(\mathcal{I^\star}))})\) can be disentangled (see \cref{thm: 1}). Essentially, this comes down to the fact that for each \( N \in \mathbf{nd}(\mathcal{I})\), the distribution \(p(\mathbf{z}, \Tilde{\mathbf{z}} \mid \text{nd}({\bm \iota}) = N)\) is an absolutely continuous measure \(\mu_N\) with non-zero mass on a submanifold of \(\mathcal{Z} \times \mathcal{Z}\) that uniquely identifies \(N\), since \(\mathbf{z}_{N} \overset{d}{=} \Tilde{\mathbf{z}}_{N}\) with respect to \(\mu_N\). Therefore we can obtain a matching of all latent blocks and their complements in the non-descendant sets of intervention targets, for any two latent causal models which produce the same observable distribution. Note that for this step, the assumption of absolute continuity of the latent distributions, together with the assumption of the smoothness of the mixing function are key.

\paragraph{Independence of interventional targets}
From \cref{sem-z2i}, we can see that given any intervention target \(S \in \mathcal{I} \), the block of post-intervention latents corresponding to \(S\) is statistically independent of the pre-intervention latents. i.e. \(\tilde{\mathbf{z}}_S \independent \mathbf{z} \mid \bm \iota = S\).
\cite{brehmer2022weakly} made use of this property to disentangle \(\mathbf{z}^\star_{S}\) for all \(S \in \mathcal{I} \), but under the restrictive assumption that \(\mathcal{I}\) consists precisely of all the \emph{atomic} intervention targets \footnote{an intervention target set \(S\) is said to be atomic if \(S = \{i\}\) for some node \(i \in V(\mathcal{G})\)}, so that no distinct intervention targets share the same non-descendant set. We remove this assumption, and instead disentangle \(\mathbf{z}^\star_{\pi(N)}\) (see \cref{thm: 2}) by making use of the fact that for all \(N \in \mathbf{nd}(\mathcal{I})\)
\begin{equation}
\label{eq: indep-of-targets}
    \tilde{\mathbf{z}}_{\pi(N)} \independent \mathbf{z} \mid \bm{nd}(\iota) = N.
\end{equation}

Loosely, the equation above translates to the fact that with respect to the distribution of \(p(\mathbf{z}, \Tilde{\mathbf{z}} \mid \text{nd}({\bm \iota}) = N)\), which we managed to separate from the other mixture components of the total paired latent distribution as a result of the previous step, \(\pi(N)\) corresponds to the maximal partition of \(\tilde{\mathbf{z}}\) that is independent from \(\mathbf{z}\). Note that the faithfulness of the causal graph is particularly important here, since we do not want to fail to take into account conditional independences which were not represented in the graph.

\subsection{DISCUSSION}
\label{subsec: discussion}
\cref{thm: 1} implies that in order to identify the latent variables corresponding to a subset of nodes \(S \in V(\mathcal{G}^\star)\) in the SCM up to abstraction, it is not necessary to have an intervention on \(S\) directly (i.e. requiring \(S \in \mathcal{I}^\star\)), which is perhaps surprising. Instead, it is sufficient to have \(S \in \sigma( \mathbf{nd}(\mathcal{I^\star}))\). 

Furthermore, this result implies that all distributional maps between ``blocks" of variables are \emph{compatible with a structural map} that abstracts \(\mathcal{G}\) to its quotient \(\mathcal{G}'\) as defined in \cref{eq: id quotient graph}. 

This is particularly relevant in cases of CRL problems where recovering the full causal graph \(\mathcal{G}\) is not possible due to lack of atomic interventions. In these scenarios, we can learn the causal structure up to an abstraction \(\mathcal{G}'\), meaning that while the causal relationships between certain latent variables corresponding to nodes in \(\mathcal{G}\) are not clear, the causal mechanisms between aggregated subsets of these variables corresponding to nodes in  \(\mathcal{G}'\) can be recovered correctly. 

Additionally, \cref{thm: 2} tells us that we can disentangle even more of the latent variables than the ones implied by \cref{thm: 1}, at the cost of disregarding the causal graph.

\section{DOWNSTREAM APPLICATIONS AND LIMITATIONS}
\label{sec: applications}

In terms of downstream tasks, our method has the usual applications for causal discovery, including causal effect estimation with respect to high-level variables; although we emphasize that in our particular problem formulation, none of the causal variables are directly observable, and therefore our work differs from the settings in classic ATE estimation or those presented in \citep{anand2023causal}.

Instead, we find that our weakly-supervised CRL setup makes assumptions that are more prevalent in recent machine learning literature, in which under unknown interventions counterfactual data pairs are indeed available, sometimes at the cost of the direct observability of causal variables. For example

\begin{itemize}
    \item In contrastive learning \citep{von2021self}, pairs of data samples before and after random augmentations or transformations, which can be viewed as interventions, are used in order to learn latent representations with causal dependencies
    \item  In problems with temporal data \citep{lippe2022citris}, we may consider sequential observations of the system as our counterfactual data. 
    \item In the field of causal interpretability, specifically with respect to the method of interchange intervention training \citep{geiger2022inducing}, we ay generate synthetic counterfactual data pairs by activation patching of neural networks.
\end{itemize}

In most machine learning applications, access to the latent causal structure can benefit generalization to out-of-distribution data. It can also serve as a foundation for interpretable and fair ML methods.

The eventual objective in CRL is having a methodology for learning the underlying latent causal model, up to some degree of abstraction. However, this is a distinct objective from having theoretical identifiability results, which demonstrate that should one succeed in learning an estimator of the model parameters which maximize the likelihood of observable distribution, then the true causal model is identified up to abstraction. 

Learning remains an important problem of its own, and we make no claim in addressing that problem. Our example in \cref{sec: exp} demonstrates a possible route for a small toy problem, and is by no mean a demonstration of a scalable method for identification of the abstract causal model, which we shall leave for future research.

\section{CONCLUSION}
\label{sec: conclusion}

We introduced a new framework for examining the identifiability of causal models up to abstraction. While previous works aiming to jointly learn the causal graph in conjunction with the latent variables have focused on fully identifying the graph up to isomorphism \citep{brehmer2022weakly, von2024nonparametric, wendong2024causal}, we show that with relaxed assumptions, we can still recover a quotient graph and additional latent blocks that can all be determined from the family of intervention targets, which are not necessarily atomic and do not have to include all nodes of the graph. We argue that this is meaningful because in some ways, the identifiable causal model abstraction to constitutes the ``real"  ground truth model given the observable distribution, by the law of parsimony.

\section*{ACKNOWLEDGMENTS}
This research was supported by CIFAR AI Chairs, NSERC Discovery, and Samsung AI Labs. Mila and Compute Canada provided computational resources. The authors would like to thank Sébastien Lachapelle for insightful discussions.

\color{black}
\bibliography{paper}     

\newpage
\section*{CHECKLIST}

 \begin{enumerate}

 \item For all models and algorithms presented, check if you include:
 \begin{enumerate}
   \item A clear description of the mathematical setting, assumptions, algorithm, and/or model. [Yes]
   \item An analysis of the properties and complexity (time, space, sample size) of any algorithm. [Not Applicable]
   \item (Optional) Anonymized source code, with specification of all dependencies, including external libraries. [No]
 \end{enumerate}

 \item For any theoretical claim, check if you include:
 \begin{enumerate}
   \item Statements of the full set of assumptions of all theoretical results. [Yes]
   \item Complete proofs of all theoretical results. [Yes]
   \item Clear explanations of any assumptions. [Yes]     
 \end{enumerate}

 \item For all figures and tables that present empirical results, check if you include:
 \begin{enumerate}
   \item The code, data, and instructions needed to reproduce the main experimental results (either in the supplemental material or as a URL). [Yes]
   \item All the training details (e.g., data splits, hyperparameters, how they were chosen). [Yes]
         \item A clear definition of the specific measure or statistics and error bars (e.g., with respect to the random seed after running experiments multiple times). [Not Applicable]
         \item A description of the computing infrastructure used. (e.g., type of GPUs, internal cluster, or cloud provider). [Not Applicable]
 \end{enumerate}

 \item If you are using existing assets (e.g., code, data, models) or curating/releasing new assets, check if you include:
 \begin{enumerate}
   \item Citations of the creator If your work uses existing assets. [Not Applicable]
   \item The license information of the assets, if applicable. [Not Applicable]
   \item New assets either in the supplemental material or as a URL, if applicable. [Not Applicable]
   \item Information about consent from data providers/curators. [Not Applicable]
   \item Discussion of sensible content if applicable, e.g., personally identifiable information or offensive content. [Not Applicable]
 \end{enumerate}

 \item If you used crowdsourcing or conducted research with human subjects, check if you include:
 \begin{enumerate}
   \item The full text of instructions given to participants and screenshots. [Not Applicable]
   \item Descriptions of potential participant risks, with links to Institutional Review Board (IRB) approvals if applicable. [Not Applicable]
   \item The estimated hourly wage paid to participants and the total amount spent on participant compensation. [Not Applicable]
 \end{enumerate}

 \end{enumerate}

\newpage
\section*{APPENDIX} 

\appendix

\section{DEFINITIONS}
\label{app: defns}

\begin{definition}[$\sigma$-algebra] 
A \emph{$\sigma$-algebra} on a set $X$ is a collection $\mathcal{F}$ of subsets of $X$ which satisfies the following properties
\begin{enumerate}
    \item Universality: \(X \in \mathcal{F}\)
    \item Closure under Complements: \(A \in \mathcal{F} \implies A^c \in \mathcal{F}\)
    \item Closure under Countable Union: \( A_n \in \mathcal{F} \quad \forall n \in \mathbb{N} \implies \bigcup_{n=1}^{\infty} A_n \in \mathcal{F}\)
\end{enumerate}
\end{definition}

\begin{definition}[Partition]
    A \emph{partition} of a set $X$ is a collection of non-empty, pairwise disjoint subsets $\{A_i\}_{i \in I}$ of $X$ such that:
\begin{enumerate}
    \item $A_i \cap A_j = \emptyset$ for all $i \neq j$ (the subsets are pairwise disjoint),
    \item $\bigcup_{i \in I} A_i = X$ (the union of all subsets covers $X$).
\end{enumerate}
\end{definition}

\begin{definition}[Group]
    A \emph{group} is a set $G$ equipped with a binary operation (often denoted by $*$) that satisfies the following four axioms:
    \begin{enumerate}
        \item Closure: $\forall a, b \in G$, $a * b \in G$
        \item Associativity: $\forall a, b, c \in G$, $(a * b) * c = a * (b * c)$
        \item Identity: $\exists e \in G$ such that $\forall a \in G$, $a * e = e * a = a$
        \item Inverse: $\forall a \in G$, $\exists a^{-1} \in G$ such that $a * a^{-1} = a^{-1} * a = e$
    \end{enumerate}
\end{definition}

\begin{definition}[Group Action]
    A \emph{group action} of a group $G$ on a set $X$ is a map $\cdot : G \times X \rightarrow X$ that satisfies the following two properties:
    \begin{enumerate}
        \item Identity: For all $x \in X$, $e \cdot x = x$, where $e$ is the identity element of $G$
        \item Compatibility: For all $g, h \in G$ and $x \in X$, $(g * h) \cdot x = g \cdot (h \cdot x)$.
    \end{enumerate}
\end{definition}

\begin{definition}[Stabilizer]
    Let \(G\) be a group acting on a set \(X\). For any element \(x \in X\), the stabilizer of \(x\), denoted by \(G_x\), is the subgroup of \(G\) consisting of all elements that fix \(x\)
    \[G_x := \{g \in G : g \cdot x = x \}\]
\end{definition}

\begin{definition}[Diffeomorphism Group]
    The diffeomorphism group on a differentiable manifold $M$, denoted by $\operatorname{Diff}(M)$, is the group of all smooth, invertible maps from $M$ to itself with smooth inverses. In other words, it consists of all bijective maps $f: M \to M$ such that both $f$ and its inverse $f^{-1}$ are differentiable. The group operation in $\operatorname{Diff}(M)$ is composition of maps. The identity element of $\operatorname{Diff}(M)$ is the identity map $\operatorname{id}_M: M \to M$. The inverse of a diffeomorphism $f \in \operatorname{Diff}(M)$ is its inverse map $f^{-1}$.
\end{definition}

\begin{definition}[Pushforward Distributions]
     The \emph{pushforward of a distribution} describes how a probability distribution is transformed under a given function. Let \((X, \mathcal{A}, \mu)\) be a measure space, where \(X\) is a set, \(\mathcal{A}\) is a $\sigma$-algebra of measurable sets on \(X\), and \(\mathbb{P}\) is a measure for a distribution on \((X, \mathcal{A})\). Let \(f: X \to Y\) be a measurable function from \(X\) to another measurable space \((Y, \mathcal{B})\). The \emph{pushforward of the measure} \(\mathbb{P}\) under \(f\), denoted by \(f_*\mathbb{P}\), is a new measure on \((Y, \mathcal{B})\) defined by:
    \[
    (f_*\mathbb{P})(B) = \mathbb{P}(f^{-1}(B)) \quad \text{for all} \quad B \in \mathcal{B}.
    \]
    In other words, the measure of a set \(B \subset Y\) under the pushforward measure \(f_*\mathbb{P}\) is the measure of its preimage \(f^{-1}(B) \subset X\) under the original measure \(\mathbb{P}\).
\end{definition}

\section{ASIDE ON DISTRIBUTIONAL ASYMMETRY}

Given random variables \(\mathbf{w}, \mathbf{w}' \in M\) and \(\mathbf{y} \in M'\) such that \(\mathbf{w} \overset{d}{=} f(\mathbf{w}', \mathbf{y})\) for some measurable function \(f: M \times M' \to M\), we are motivated by the question: does the statistical independence of \(\mathbf{w}\) and \(\mathbf{y}\) imply the functional independence of \(\mathbf{w}\) and \(\mathbf{y}\)? More formally, we want to know if the following holds
\begin{equation}
\label{eq: stat indep implies fn indep}
   \mathbf{w} \independent \mathbf{y} \implies \exists \hat{f} : M \to M \text{ s.t. } f(\cdot , y) = \hat{f}  \quad \forall y \in M'
\end{equation}
At first glance, the answer might seem to be yes. But suppose that \(\mathbf{w}\) has a standard Gaussian distribution on \(\mathbb R^2\), \(\mathbf{y}\) has a uniform distribution over \([0, \pi]\) and \(f(w, y) := R_{y}(w) \), where \(R_{y}\) denotes a clockwise rotation about the origin in \(\mathbb R^2\). Then it is easy to see that the distribution \(p(\mathbf{w} \mid \mathbf{y} = y)\) is a 2D standard Gaussian for all \(y\), so \(\mathbf{w} \independent \mathbf{y}\), but certainly \(f\) is not constant with respect to its second argument.

One way to guarantee that statistical independence implies function independence is by assuming that \(f\) is smooth and that \(p(\mathbf{w}')\) cannot be preserved by a ``smoothly varying" family of diffeomorphisms, which we will show below.

\begin{proposition}
\label{prop: asymmetry plus stat indep implies fn indep}
    Suppose the stabilizer of the distribution \(\mathbb{P}_{\mathbf{w}'}\) is totally disconnected in the diffeomorphism group $\operatorname{Diff}(M)$. Then the condition in \cref{eq: stat indep implies fn indep} holds. 
\end{proposition}

\begin{proof}
    First, note that the diffeomorphism group $\operatorname{Diff}(M)$ can act on the space of probability distributions on $M$ in a natural way. For any diffeomorphism $f \in \operatorname{Diff}(M)$ and any probability distribution $\mathbb{P}$ on $M$, the pushforward of $\mathbb{P}$ by $f$, denoted by $f_{*}\mathbb{P}$, is a new probability distribution on $M$. We can thus easily check that the group action \(f \cdot \mathbb{P} := f_{*}\mathbb{P}\) is well-defined.

    Next, for any metric \(d\) on \(M\) we can equip \(\operatorname{Diff}(M)\) with the following metric, where for \(f, g \in \operatorname{Diff}(M)\) we have \(d_{\infty}(f, g) := \sup_{x \in M}d(f(x), g(x))\). A set \(S \subseteq \operatorname{Diff}(M)\) is said to be \emph{totally disconnected} if there is no \(\epsilon > 0\) and \(f \in \operatorname{Diff}(M)\) such that \(B(f, \epsilon) = \{g \in \operatorname{Diff}(M): d_{\infty}(f, g) < \epsilon\} \subseteq S\). Intuitively, this means there is no smoothly varying family of diffeomorphisms in \(S\), however small.
    
    Finally, consider the set of diffeomorphisms \(S = \{f(\cdot, y) : y \in M' \}  \subseteq \operatorname{Diff}(M)\). By the smoothness of \(f\) we can see that \(S\) is connected in \(\operatorname{Diff}(M)\). Since the stabilizer of the distribution \(\mathbb{P}_{\mathbf{w}'}\) is totally disconnected in the diffeomorphism group $\operatorname{Diff}(M)$, \(\mathbf{w} \independent \mathbf{y}\) implies that 
    \begin{equation}
        f(\cdot, y)_* \mathbb{P}_{\mathbf{w'}} = \mathbb{P}_{\mathbf{w}} \quad \forall y \in M'
    \end{equation}
    which means that \(S\) is contained a left coset of the stabilzer of \(\mathbb{P}_{\mathbf{w}'}\), and therefore must be totally disconnected. But since \(S\) is connected this means that it is a singleton set, and hence there exists \(\hat{f} \in \operatorname{Diff}(M)\) such that \(f(\cdot, y) = \hat{f}\) for all \(y \in M'\).
\end{proof}

\section{PROOFS}
\label{app: proofs}

\textbf{Theorem 3.1} \label{thm1} Under the following assumptions on the hypothesis class of parameters \(\Theta\), latent causal model with parameters \(\theta^\star \in \Theta\) is identifiable up to a causal model abstraction \({\theta}\) with directed acyclic causal graph \({\mathcal{G}} = \mathcal{G}^\star / \mathcal{P}({\sigma( \mathbf{nd}(\mathcal{I^\star}))})\)

\begin{enumerate}
    \item \emph{Faithfulness of the causal graph}: Let \(\mathcal{G}\) be a perfect map for the distribution of \(p(\mathbf{z})\), meaning that it encapsulates all the conditional independences.
    \item  \label{assump: 2} \emph{Absolute continuity of latent distributions}: Let \(\mathcal{E}_i \cong \tilde{\mathcal{E}}_i \cong \mathcal{Z}_i\), let \(f_i\), \(f_i^{(S)}\) be continuously differentiable for all \(i\) and \(S\), and let \(p_{\bm{\varepsilon}}\) and \(p_{\tilde{\bm{\varepsilon}}}\) be absolutely continuous.
    \item \label{assump: 3} \emph{Smoothness of mixing function}:  Let \(g\) be a diffeomorphism.
\end{enumerate}

\begin{proof}

\label{proof: nd thm}

\vspace{2em}

Suppose we have \(\theta' \in \Theta\) such that \(p_{\theta'}(\mathbf{x}, \tilde{\mathbf{x}}) = p_{\theta^\star}(\mathbf{x}, \tilde{\mathbf{x}})\). Denote the measures representing the distributions of \(p_{\theta'}(\mathbf{z}, \tilde{\mathbf{z}})\) and \(p_{\theta^\star}(\mathbf{z}, \tilde{\mathbf{z}})\) as \(\mu\) and \(\nu\) respectively. Denote their marginals \(p_{\theta'}(\mathbf{z})\) and \(p_{\theta^\star}(\mathbf{z})\) as \(\mu_0\) and \(\nu_0\) respectively. Let \(h:= (g')^{-1} \circ g^\star\), and write \(h^{\otimes 2}: (z, \Tilde{z}) \mapsto (h(z), h(\Tilde{z}))\). Then by definition of the data generating process

\begin{equation}
\label{eq: paired latent equiv}
    \mu = h^{\otimes 2}_* \nu
\end{equation}

\textbf{Finite Mixtures conditioned on Non-Descendants}
We can decompose these distributions into finite mixtures, so that we have 

\begin{align}
    \mu &= \sum_{M \in \mathbf{nd}(\mathcal{I'})} \alpha_{M} \mu_{M} \\
    \nu &= \sum_{N  \in \mathbf{nd}(\mathcal{I}^\star)} \beta_{N} \nu_N
\end{align}

where  \(\mu_M\) and \(\nu_N\) denote the distributions \(p_{\theta'}(\mathbf{z}, \tilde{\mathbf{z}} \mid \mathbf{nd}(\bm \iota) = M)\) and \(p_{\theta^\star}(\mathbf{z}, \tilde{\mathbf{z}} \mid  \mathbf{nd}(\bm \iota) = N)\), and \(\alpha_M\) and \(\beta_N\) denote the positive constants \(p_{\theta'}(\mathbf{nd}(\bm \iota) = M)\) and \(p_{\theta^\star}(\mathbf{nd}(\bm \iota) = N)\).

\textbf{Distributions of Latent Differences}
 By a slight abuse of notation define the following operator on both  \(\mathcal{Z}^\star \times \mathcal{Z}^\star\) and \(\mathcal{Z}' \times \mathcal{Z}'\).

\begin{equation}
    \Delta: (\mathbf{z} , \tilde{\mathbf{z}}) \mapsto \mathbf{z} - \tilde{\mathbf{z}}
\end{equation}

For any \(N \in \mathbf{nd}(\mathcal{I^\star})\) note that the support of \(\Delta_* \nu_N\) is restricted to the subspace \(\{\bm{0}_N\} \times \mathcal{Z}^\star_{N^c}\), and moreover \(N\) is the maximal subset \(S \subseteq V(\mathcal{G})\) such that the support of \(\Delta_* \nu_N\) can be restricted to the subspace \(\{\bm{0}_S\} \times \mathcal{Z}^\star_{S^c}\).

So for distinct  \(N_1, N_2 \in \mathbf{nd}(\mathcal{I^\star})\), \(N_1 \cap N_2\) is the maximal subset \(S \subseteq V(\mathcal{G}^\star)\) such that the union of the supports of \(\Delta_* \nu_{N_1}\) and \(\Delta_* \nu_{N_2}\) can be restricted to the subspace \(\{\bm{0}_S\} \times \mathcal{Z}^\star_{S^c}\), and \(N_1 \cup N_2\) is the maximal subset \(S \subseteq V(\mathcal{G}^\star)\) such that the intersection of the supports of \(\Delta_* \nu_{N_1}\) and \(\Delta_* \nu_{N_2}\) can be restricted to the subspace \(\{\bm{0}_S\} \times \mathcal{Z}^\star_{S^c}\).

Similarly, for distinct  \(M_1, M_2 \in \mathbf{nd}(\mathcal{I'})\), \(M_1 \cap M_2\) is the maximal subset \(S \subseteq V(\mathcal{G}')\) such that the union of the supports of \(\Delta_* \mu_{M_1}\) and \(\Delta_* \mu_{M_2}\) can be restricted to the subspace \(\{\bm{0}_S\} \times \mathcal{Z}'_{S^c}\), and \(M_1 \cup M_2\) is the maximal subset \(S \subseteq V(\mathcal{G}')\) such that the intersection of the supports of \(\Delta_* \mu_{M_1}\) and \(\Delta_* \mu_{M_2}\) can be restricted to the subspace \(\{\bm{0}_S\} \times \mathcal{Z}'_{S^c}\).

\textbf{Separating Non-Descendant Mixtures}
For all \(N \in  \mathbf{nd}(\mathcal{I^\star})\) we can let \(\varphi(N)\) be the maximal subset \(M \in \mathbf{nd}(\mathcal{I}')\) such that  \(h(\operatorname{supp}(\nu_N)) \subseteq \{\bm{0}_M\} \times \mathcal{Z}'_{M^c}\), so that this defines a bijection 

\begin{equation}
    \varphi : \mathbf{nd}(\mathcal{I^\star}) \to \mathbf{nd}(\mathcal{I}')
\end{equation}

such that for all \(N \in  \mathbf{nd}(\mathcal{I^\star})\)

\begin{align}
    &\alpha_{\varphi(N)} = \beta_N \\
    &\mu_{\varphi(N)} = h^{\otimes 2}_* \nu_N \label{eq: sep mixtures}
\end{align}

\textbf{Disentangling Non-Descendant Sets}
The projection of \(\nu\) onto \(\Delta^{-1}(\{\bm{0}_{N}\} \times \mathcal{Z}^\star_{N^c})\) consists only of components \(\nu_{N'}\) such that \(N' \supseteq N\). So under \(\nu\), given \(\mathbf{z}_N = \tilde{\mathbf{z}}_N\), the joint distribution of the latent variables \(h(\mathbf{z})\) and \(h(\tilde{\mathbf{z}})\) is represented by a corresponding mixture of components \(\mu_{\varphi(N')}\) where \(\varphi(N') \supseteq \varphi(N)\), so the support of each \(\Delta_* \mu_{\varphi(N')}\) is restricted to \( \{\bm{0}_{\varphi(N)}\} \times \mathcal{Z}'_{\varphi(N)^c}\). Therefore \(\mathbf{z}_N = \tilde{\mathbf{z}}_N\) almost surely implies \(h(\mathbf{z})_{\varphi(N)} = h(\tilde{\mathbf{z}})_{\varphi(N)}\) almost surely, meaning that \(h(\mathbf{z})_{\varphi(N)}\) is conditionally independent of \(\mathbf{z}_{N^c}\) given \(\mathbf{z}_N\). Similarly, we can show that \(\mathbf{z}_N\) is conditionally independent of \(h(\mathbf{z})_{\varphi(N)^c}\) given \(h(\mathbf{z})_{\varphi(N)}\). Hence there exists \(h_N\) such that 

\begin{equation}
   \mathbf{z}_{\varphi(N)} \overset{d}{=}  h_N( \mathbf{z}^\star_N)
\end{equation}

\textbf{Disentangling Complements of Non-Descendants}
The projection of \(\nu\) onto \(\Delta^{-1}(\{\bm{0}_{N^c}\} \times \mathcal{Z}^\star_{N})\) only contains components \(\nu_{N'}\) if \(N' \supseteq N^c\).
Thus \(\mathbf{z}_{N^c} = \tilde{\mathbf{z}}_{N^c}\) almost surely implies \(h(\mathbf{z})_{M} = h(\tilde{\mathbf{z}})_{M}\) almost surely for all \(M \in \mathbf{nd}(\mathcal{I}')\) with \(\varphi^{-1}(M) \supseteq N^c\), so \(h(\mathbf{z})_{\varphi(N)^c} = h(\tilde{\mathbf{z}})_{\varphi(N)^c}\) almost surely. Therefore \(h(\mathbf{z})_{\varphi(N)^c}\) is conditionally independent of \(\mathbf{z}_{N}\) given \(\mathbf{z}_{N^c}\). Similarly, we can show that \(\mathbf{z}_{N^c}\) is conditionally independent of \(h(\mathbf{z})_{\varphi(N)}\) given \(h(\mathbf{z})_{\varphi(N)^c}\). Hence there exists \(h_{N^c}\) such that 

\begin{equation}
   \mathbf{z}_{\varphi(N)^c} \overset{d}{=}  h_{N^c}( \mathbf{z}^\star_{N^c})
\end{equation}

\textbf{Disentangling Intersections}
If there are functions \(h_A\) and \(h_B\) identifying \( \mathbf{z}^\star_A\) and \( \mathbf{z}^\star_B\) then naturally their marginals would agree on \(\mathcal{Z}^\star_{A \cap B}\) such that they define a new function \(h_{A \cap B}\) that identifies  \( \mathbf{z}^\star_{A \cap B}\).

\textbf{Identifying Quotient Graph}
Given that we have shown that we can identify all latents corresponding to complements of non-descendant sets as well as intersections, we can extend the \(\varphi\) such that for any \(A, B \in \sigma( \mathbf{nd}(\mathcal{I^\star}))\) we have \(\varphi(A^c) = \varphi(A)^c\) and \(\varphi(A \cap B) = \varphi(A) \cap \varphi(B)\), and note that this is a bijection from \(\sigma( \mathbf{nd}(\mathcal{I^\star}))\) to \(\sigma( \mathbf{nd}(\mathcal{I'}))\). Furthermore, using \(\mathcal{P}^\star\) and \(\mathcal{P}'\) to denote \(\mathcal{P}(\sigma(\mathbf{nd}(\mathcal{I^\star})))\) and \(\mathcal{P}( \sigma(\mathbf{nd}(\mathcal{I}')))\) respectively, we can now restrict \(\varphi\) to a bijection

\begin{equation}
    \varphi : \mathcal{P}^\star \to \mathcal{P}'
\end{equation}

This matching of partitions constitutes a graph isomorphism between the quotient graphs  \(\mathcal{G}^\star / \mathcal{P}^\star \) and  \(\mathcal{G}' / \mathcal{P}' \), which is equivalent to a graph epimorphism from \(\mathcal{G}\) to \(\mathcal{G}^\star / \mathcal{P}^\star\). To show this, note that any edge in the quotient graph \(\mathcal{G}^\star / \mathcal{P}^\star \) from a source block \(A\) to a terminal block \(B\) is the result of causal dependency between some vertices \(i \in A\) and \(j \in B\), by faithfulness of the original graph. So since \(\mathbf{z}_j \not \independent \mathbf{z}_i \mid \mathbf{z}_{\operatorname{Pa}(j)}\), we must have \(\mathbf{z}_B \not \independent \mathbf{z}_A \mid \mathbf{z}_{\operatorname{Pa}(B)}\), which means \(h_B(\mathbf{z}_B) \not \independent h_A(\mathbf{z}_A) \mid h_{\operatorname{Pa}(B)}(\mathbf{z}_{\operatorname{Pa}(B)})\). Therefore there is an edge from \(h(A)\) to \(h(B)\) in \(\mathcal{G}' / \mathcal{P}' \). The same of course holds for \(\varphi^{-1}\) and \(h^{-1}\).

Finally, we can quickly note that \(\hat{G} = \mathcal{G}^\star / \mathcal{P}^\star\) is indeed acyclic, since any cycle in \(\hat{\mathcal{G}}\) implies that there exists a directed edge from the complement \(V(\mathcal{G}^\star) \setminus N \) of a non-descendant set \(N \in  \mathbf{nd}(\mathcal{I}^\star)\) to \(N\) itself, which violates the definition of a non-descendant set.
    
\end{proof}

\begin{theorem}
\label{thm2}
    For any \(N \in \mathbf{nd}(\mathcal{I}^\star)\) denote the intersection of all intervention targets with \(N\) as their non-descendant set as \[\pi(N) := \bigcap \{S \in \mathcal{I}: \text{nd}(S) = N\}\] 
    Now provided that \(\pi(N)\) is a singleton set \(\{i\}\) and \(\mathcal{Z}^\star_{i} \cong \mathbb R\), then we can identify the latent \(\mathbf{z}^\star_{\pi(N)}\) up to disentanglement, meaning that for all \(\theta \in \Theta\)
    \begin{equation}
         p_{\theta}(\mathbf{x},  \Tilde{\mathbf{x}}) = p_{\theta^\star}(\mathbf{x}, \Tilde{\mathbf{x}}) \implies   {\theta} \sim_{L} {\theta^\star} \quad \text{wrt } \mathcal{Z}^\star_{i}
    \end{equation}
\end{theorem}

\begin{proof}

To show that we can additionally identify \(\pi(N)\), note that for any \(N \in \mathbf{nd}(\mathcal{I}^\star)\), by definition of the data generating process, there exists some deterministic function \(F_{\theta^\star}\) such that 

\begin{equation}
    \tilde{\mathbf{z}}^\star_{\pi(N)^c} = F_{\theta^\star}(\mathbf{z}^\star,  \tilde{\mathbf{z}}_{\pi(N)})
\end{equation}

Thus we can deduce

\begin{align}
    \tilde{\mathbf{z}}_{\pi(N)}  
    &\overset{d}{=} h(\tilde{\mathbf{z}}^\star) \\
    &= h(\tilde{\mathbf{z}}^\star_{\pi(N)} +  F_{\theta^\star}(\mathbf{z}^\star,  \tilde{\mathbf{z}}^\star_{\pi(N)}))_{\pi(N)}  \\
    &= h(\tilde{\mathbf{z}}^\star_{\pi(N)} +  F_{\theta^\star}(h^{-1}(\mathbf{z}),  \tilde{\mathbf{z}}^\star_{\pi(N)}))_{\pi(N)} 
\end{align}

Note also that the latent distribution \((\mathbf{z}, \tilde{\mathbf{z}})\) under \(\theta\) satisfies \(\tilde{\mathbf{z}}_{\pi(N)} \independent \mathbf{z} \mid \operatorname{nd}(\bm \iota) = N\). Therefore by \cref{prop: asymmetry plus stat indep implies fn indep}, there exists \(h_{\pi(N)}\) such that \(\tilde{\mathbf{z}}_{\pi(N)}  \overset{d}{=} h_{\pi(N)}(\tilde{\mathbf{z}}^\star_{\pi(N)}) \), since \(\tilde{\mathbf{z}}^\star_{\pi(N)})\) is a random variable with domain on \(\mathbb{R}\), so the stabilizer of its distribution contains at most two elements -- the identity and a reflection about a point on \(\mathbb{R}\) -- and thus must be totally disconnectd.

\end{proof}

\section{EXPERIMENTS}
\label{sec: exp}

\paragraph{SETUP}
We generate synthetic data for linear Gaussian models, where the parameter space \(\Theta\) is defined as follows. For any directed acyclic graph \(\mathcal{G}\) with \(n\) nodes, we let each of the nodes \(i \in V(\mathcal{G})\) be equipped with the latent space \(\mathbb{R}\). We let the distribution of the exogenous variables be a standard isotropic unit Gaussian \(\bm \varepsilon \sim \mathcal{N}(\bm 0, \mathbf{I}_n)\) and furthermore define \(\mathbf{z}_i = \sum_{j \in Pa_{\mathcal{G}}(i)} a_{ij} \mathbf{z}_j + \bm \varepsilon_i \), where the coefficients \(a_{ij}\) are sampled from a Gaussian mixture model that has two equal components with means ±1 and standard deviation 0.25. For the intervention parameters, we sample the new exogenous variables \(\tilde{\bm \varepsilon}_S\) from unit Gaussians again and let \(\tilde{f}_i^{(S)}\) be the identity for all subsets \(S \subseteq V(\mathcal{G})\) and \(i \in S\). We then uniformly sample a rotation matrix \(Q \in SO(n)\) with respect to the Haar measure for our mixing function \(g\).

\paragraph{RESULTS}
We maximize the likelihood of a set of latent causal parameters \(\theta\) over the parameter space \(\Theta\) for linear Gaussian models as described above via gradient descent, with respect to the observed data. We validate our theory by showing that for parameters with sufficiently high likelihood, the learned encoder \(Q^T\) inverts the ground truth mixing function \(Q^\star\) up to the required level of abstraction, meaning that in particular, as a result of \cref{thm: 1}, \(Q^T Q^\star\) is approximately a block diagonal matrix where each of the blocks corresponds to an element in \(\mathcal{P}({\sigma( \mathbf{nd}(\mathcal{I^\star}))})\).

\end{document}